# Negation Detection for Clinical Text Mining in Russian

Anastasia FUNKNER[a, 1], Ksenia BALABAEVA[a,] and Sergey KOVALCHUK[a]

[a] *ITMO University, Saint Petersburg, Russia*

**Abstract.** Developing predictive modeling in medicine requires additional features from unstructured clinical texts. In Russia, there are no instruments for natural language processing to cope with problems of medical records. This paper is devoted to a module of negation detection. The corpus-free machine learning method is based on gradient boosting classifier is used to detect whether a disease is denied, not mentioned or presented in the text. The detector classifies negations for five diseases and shows average F-score from 0.81 to 0.93. The benefits of negation detection have been demonstrated by predicting the presence of surgery for patients with the acute coronary syndrome.

## 1. Introduction

An electronic medical record (EMR) contains records in natural language, for example, life and disease anamnesis, protocols of surgeries, examination or discharge reports. Developing predictive modeling in medicine requires additional features, and not only strictly structured ones, such as blood tests, patient metrics, etc. For the English language, there are many tools for labeling text, extracting entities, disease's cases, temporal and negation detection (UMLS, UIMA, IBM Watson, Apache Ruta, etc.) [1–4]. However, most of the tools are now impossible to adapt to the Russian language because a word corpus of medical texts has not yet been compiled. There is a small one based on 120 records with labeled diseases and their attributes (complications, severity, treatment, etc.) [5,6]. But this corpus is compiled on the records of one medical centre and for patients with a limited set of diseases, and therefore it is difficult to use it for automatic text labeling of other medical centres. Thus, our team aims to develop a system of four modules to solve problems of misprints in medical terms, negation, temporality, and experiencer detection (see Figure 1). This paper is devoted to a module of negation detection.

It is accepted that any condition or disease is sought by the occurrence of synonyms-terms in medical history, anamnesis, etc. However, the presence of such a term may also indicate that the patient denies the disease. In Russia, healthcare providers often ask patients after admission to the hospital about strokes, heart attacks, and high blood pressure in the past, as this may affect the course of treatment. As a result, in disease anamnesis, phrases like "MI in the past denies" (Myocardial Infarction), "there was no increase in blood pressure", "without a history of hypertension", "diabetes denies" can be often found.

---

[1] Corresponding Author, ITMO University, 49 Kronverksky Pr., St. Petersburg, 197101, Russia; E-mail: funkner.anastasia@itmo.ru, funkner.anastasia@gmail.com.

The above-mentioned problem is called negation detection in texts. One of the first algorithms NegEx was proposed by Chapman et al. in 2001 and was based on rules [7]. Subsequently, syntax-based methods for the English language were developed [8,9]. Goryachev et al. implemented four methods for detecting negations and compared them using three sets of clinical records in English [10]. NegEx showed the best accuracy [7]. Thus, most of the approaches found for identifying negations are developed with grammatical rules for the English language.

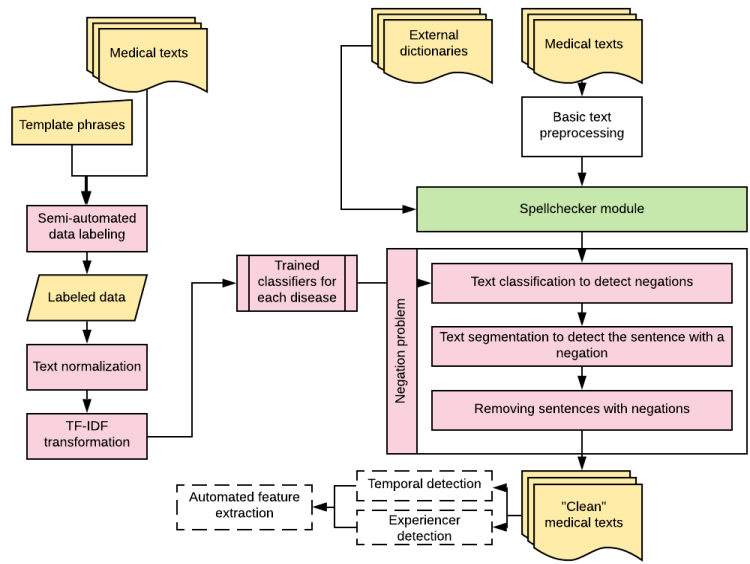

**Figure 1**. Modules of spelling correction and negation detection. Pink blocks indicate methods which are described in this paper. Green blocks are developed and implemented by colleagues. Dashed blocks indicate methods that are just being developed.

## 2. Method

In this research, we used a set of anonymized 3434 EMRs of patients with the acute coronary syndrome (ACS) who admitted to Almazov National Medical Research Centre (Almazov Centre) during 2010-2015. Disease anamneses are one of the most unstructured records (free text without any tags; each physician writes as he/she wants), and therefore we used them to demonstrate our approach.

Our approach to detect negations is machine learning based on multi-class classification. In Figure 1 the main steps are presented on how to train a negation detector and use for new data sets. Firstly, clinical texts should be annotated with 3 labels for each specific disease: '-1' means 'the disease is denied', '0' means 'the disease is not mentioned in the text' and '1' means 'patient has this disease'. We suppose to annotate the whole anamnesis with these labels because the classifier will be able to learn an additional context which follows the disease. The two problems are possible: the disease can relate to a patient's family members (label with 0) and there can be two descriptions about patient's condition in different moments of her or his life (label with 1). We use semi-automated labeling with phrase search because the provided anamneses can be written in a similar format by the same doctor and it is possible to find repeated phrases.

Secondly, annotated anamnesis is normalized with pymorphy2 module which can be used for Russian and Ukraine [11]. Term frequency – Inverse document frequency (TF-IDF) transformation makes commonly used words less weighted in word corpus (prepositions, pronounce, etc.). TF-IDF vectors are used to train, tune and chose the best classifier. In consequence of cross-validation experiments, based on decision trees a gradient boosting classifier shows the best result with F-score. Thus, we trained five classifiers for detecting arterial hypertension (AH), myocardial infarction (MI), stroke, diabetes mellitus (DM) and angina pectoris (AP) (see Table 1). Using negation classifiers is crucial for stroke, MI and AH. For MI and AH, the classifiers learn context and help find additional cases of these conditions. Table 2 includes the most import phrases for detecting MI which are disease terms and possible treatment (surgery and medications).

When the negations are detected, each sentence of the anamnesis is labeled as containing or not containing negation using a logistic loss [12]. Sentences or their parts with negations are removed from anamnesis so that these texts can be used to build other models and consider only the patient's existing conditions, for instance, for topic modeling.

**Table 1.** The comparative results for negation classifiers (F1 – F-score). False positives are anamneses which were detected with '1' and '-1' using a term search and a negation classifier respectively. False negatives were detected with '0' (terms not found in a text) and '1' with the same methods.

| Disease | Number of annotated anamneses | F1 for '-1' labeled | F1 for '0' labeled | F1 for '1' labeled | Macro average F1 | Total accuracy | Number of false positives (%) | Number of false negatives (%) |
|---|---|---|---|---|---|---|---|---|
| **stroke** | 648 | 0.96 | 0.96 | 0.78 | 0.9 | 0.93 | **0.15** | 0.00 |
| **MI** | 897 | 0.95 | 0.5 | 0.97 | 0.81 | 0.96 | 0.07 | **0.15** |
| **AH** | 1167 | 0.88 | 0.59 | 0.96 | 0.81 | 0.93 | 0.08 | **0.20** |
| **DM** | 239 | 0.95 | 0.93 | 0.91 | 0.93 | 0.93 | 0.02 | 0.00 |
| **AP** | 864 | 0.67 | 0.91 | 0.98 | 0.85 | 0.95 | 0.02 | 0.01 |

**Table 2.** Feature importance (FI) of words and phrases which are used to define the presence of disease in an anamnesis.

| Russian phrase | English phrase | FI | Class | Comments |
|---|---|---|---|---|
| оим | AMI | 0.052 | 1 | Abbreviation: Acute Myocardial Infarction |
| ибс оим | IHD AMI | 0.042 | 1 | Abbreviation: Ischemic Heart Disease. IHD includes myocardial infarction. |
| миокард отрицать | myocardial deny | 0.041 | -1 | Patient denies MI |
| отрицать | deny | 0.035 | -1 | Patient denies an acute condition |
| миокард ранее | myocardial early | 0.027 | 1 | Indicates myocardial problems previously |
| нижний | low | 0.022 | 1 | Indicates the localization of AMI |
| провести стентирование | perform stenting | 0.016 | 1 | Stenting is a surgery performed after AMI. |
| ознакомить | provide | 0.015 | 0 | Indicates anamnesis without a medical history |

| передний | anterior | 0.014 | 1 | Indicates the localization of AMI |
| анальгин трамадол | analgin, tramadol | 0.013 | 1 | Medicines that are used to treat AMI |

## 3. Results

To evaluate how the application of negation detection module affects predictive model performance we conduct experiments on 2 tasks: clinical episode outcome prediction and prediction of surgery for the patients suffering from ACS [13]. Usually, all information medics have about a patient by the time of arrival is clinical anamnesis in the natural text format. Therefore, we use it as input data to train machine learning models.

During experiments all the models parameters remain constant (for XGBoost: num. estimators = 500, max depth = 100, learning rate = 0.1, objective = binary logistic; for Random Forest: max depth = 100, num estimators = 500, criterion = 'gini', min. samples split = 2, min. samples leaf = 1; for k-nearest neighbors model (KNN): num. neighbours = 10, weights = 'uniform'). The only change is in the input data for the fixed feature set: we use diagnosis features collected from clinical texts using regular expressions without negation detector and with the help of negation detector.

We evaluate the results on 33% test sample using the F1 score as a quality metric. The results are performed in Table 3.

**Table 3.** Evaluation of the negation detection (ND) performance on predictive tasks.

| Task | Model | F1 without ND Text features | F1 with ND Text features | F1 without ND Text + other features | F1 with ND Text + other features |
|---|---|---|---|---|---|
| **Surgery prediction** | XGBoost | 0.3755 | 0.5234 | 0.6941 | 0.6992 |
|  | Random Forest | 0.3948 | 0.7110 | 0.7038 | 0.7128 |
|  | KNN | 0.5091 | 0.5234 | 0.5851 | 0.5870 |
| **Outcome prediction** | XGBoost | 0.3234 | 0.3233 | 0.5208 | 0.5113 |
|  | Random Forest | 0.3234 | 0.4229 | 0.4402 | 0.4569 |
|  | KNN | 0.3234 | 0.3233 | 0.3428 | 0.3428 |

## 4. Discussion

According to the experiment results, we may conclude that the use of negation detector significantly improves the performances of XGBoost, Random Forest and KNN on the task of surgery prediction based on only text features. On the task of the outcome prediction, only Random forest improves. All other algorithms' results do not change. One of the reasons could be the need for additional information for the outcome prediction to improve since on current feature sample models predict the constant class for all instances and can't properly distinguish classes. For this reason, we conduct additional experiments and add other numerical features based on the first lab test the patients had. Lab tests for models are more significant than texts, therefore, the effect from the negation detection application is not such significant in this experimental set.

## 5. Conclusion

We demonstrated the procedure for training and using the negation detector on the example of disease anamnesis of patients with ACS. For five treatment-relevant diseases and conditions, the negation detector showed average F-score from 0.81 to 0.93 on the test sample. Also, the benefits of the negation detection were demonstrated with predicting the presence of surgery for patients with ACS. The developed and implemented module for detecting and removing negations is part of the application for processing clinical text. In the future, we plan to develop additional modules and make this application an indispensable part of data preparation and feature extraction for any predictive modeling.

**Acknowledgements**

This research is financially supported by The Russian Science Foundation, Agreement #19-11-00326.